\documentclass{article}

    \PassOptionsToPackage{numbers, compress}{natbib}

\usepackage[preprint]{neurips_2020}

\usepackage[utf8]{inputenc} 
\usepackage[T1]{fontenc}    
\usepackage{hyperref}       
\usepackage{url}            
\usepackage{booktabs}       
\usepackage{amsfonts}       
\usepackage{nicefrac}       
\usepackage{microtype}      

\title{Learning Multiclass Classifier Under Noisy Bandit Feedback}

\author{%
  Mudit Agarwal\\
  Machine Learning Lab\\
  IIIT Hyderabad \\
  \texttt{mudit.agarwal@research.iiit.ac.in} \\
   \And
  Naresh Manwani  \\
  Machine Learning Lab\\
  IIIT Hyderabad \\
  \texttt{naresh.manwani@iiit.ac.in} \\
}

\usepackage{soul}
\usepackage[small]{caption}
\usepackage{graphicx}
\usepackage{algorithm}
\usepackage{algorithmic}
\usepackage{amssymb}
\usepackage{amsmath}
\usepackage{amsthm}
\usepackage{subfig}
\usepackage{enumitem}
\usepackage{multirow}
\usepackage{appendix}
\usepackage{paracol}
\usepackage{multicol}

\newtheorem{theorem}{Theorem}
\newtheorem{lemma}[theorem]{Lemma}
\newtheorem{corollary}{Corollary}[theorem]

\usepackage{times}
\usepackage{graphicx}
\usepackage{soul}
\usepackage{algorithm}
\usepackage{algorithmic}
\usepackage{amssymb}
\usepackage{amsmath}
\usepackage{amsthm}
\usepackage{subfig}
\usepackage{enumitem}
\usepackage{multirow}
\usepackage{appendix}
\usepackage{paracol}
\usepackage{multicol}
\usepackage[utf8]{inputenc} 
\usepackage{hyperref}       
\usepackage{booktabs}       
\usepackage{amsfonts}       
\usepackage{nicefrac}       
\usepackage{microtype}      

\newcommand{\norm}[1]{\left\lVert#1\right\rVert}

\begin{document}

\maketitle

\def \E {\mathbb{E}}
\def \I {\mathbb{I}}
\def \f {f_{\rho}}
\def \ri {\rho_{1}}
\def \rz {\rho_{0}}
\def \epz{\epsilon_{0}}
\def \epi{\epsilon_{1}}
\maketitle              

\begin{abstract}
This paper addresses the problem of multiclass classification with corrupted or noisy bandit feedback. In this setting, the learner may not receive true feedback. Instead, it receives feedback that has been flipped with some non-zero probability. We propose a novel approach to deal with noisy bandit feedback based on the unbiased estimator technique. We further offer a method that can efficiently estimate the noise rates, thus providing an end-to-end framework. The proposed algorithm enjoys a mistake bound of the order of $O(\sqrt{T})$ in the high noise case and of the order of $O(T^{\nicefrac{2}{3}})$ in the worst case. We show our approach's effectiveness using extensive experiments on several benchmark datasets.
\end{abstract}
\section{Introduction}
In machine learning, multiclass classification is of particular  interest due to its widespread application in several domains such as  digit-recognition \cite{ma2015effective}, text classification \cite{mccallum1999multi} and  recommender systems \cite{kakade2008efficient}. Some of the well-known batch learning approaches for multiclass classification are discussed in \cite{hsu2002comparison,anthony2009neural,bishop1995neural,ou2007multi}. An extension of Perceptron \cite{rosenblatt1958perceptron} to the multiclass setting was first proposed in \cite{duda1973pattern}, which was later modified by \cite{kakade2008efficient} to deal with bandit feedback setting.
Unlike the full information setting, the bandit setting's learner receives only partial feedback, indicating whether the predicted label is correct or incorrect, popularly known as bandit feedback. The learner's ability to learn a correct hypothesis under bandit feedback finds several web-based applications, such as sponsored advertising on web pages and recommender systems as mentioned by \cite{kakade2008efficient}. In the typical setting of the recommender system, when a user makes a query to the system, then the user is presented with a suggestion based on the past browsing history; finally, the user responds to the suggestion, either positively (clicking it) or negatively (not clicking it). However, the system does not know the behavior of the user if presented with other suggestions.

 Banditron \cite{kakade2008efficient} uses an exploitation-exploration scheme proposed in \cite{Auer:2003}. When it updates, it replaces the gradient of the loss function with an unbiased estimator of the gradient. When the data is linearly separable, the expected number of mistakes made by Banditron is shown to be $O(\sqrt{T})$. In the general case, the expected number of mistakes of Banditron is $O\left(T^{\nicefrac{2}{3}}\right)$. 
Another bandit algorithm, named Newtron \cite{NIPS2011_fde9264c}, is based on the online Newton method. It uses a  strongly convex objective function (adding regularization term with the loss function) and Follow-The-Regularized-Leader (FTRL) strategy to achieve $O(\log T)$ regret bound in the best case and $O\left(T^{\nicefrac{2}{3}}\right)$ regret bound in the worst case. Second-order Perceptron is also extended in bandit feedback setting by Crammer, and Gentile \cite{bar}. It uses
upper-confidence bounds (UCB) \cite{Auer:2002} based approach to handle exploration-exploitation and achieves regret bound of $O\left(\sqrt{T}\log(T)
\right)$ Beygelzimer et al. \cite{bmlc} proposed efficient algorithms under bandit feedback when the data is linearly separable by a margin of $\gamma$. They show that their algorithm achieves a near-optimal bound of $O\left(\nicefrac{K}{\gamma}\right)$ under strong linear separability condition \cite{bmlc}. 

In all the above approaches, it is assumed that the user has provided correct bandit feedback. There are many practical situations where the bandit feedback can become noisy too. It means that the feedback that the predicted label is the same as the actual label can be wrong with some non-zero probability. Consider the following examples of noisy bandit feedback. 
In the recommendation system, there are few cases in which a user may accidentally click (positive feedback) the recommended ad. In this case, the true feedback should be negative (no clicks). However, instead of negative, the recommender system receives positive feedback. Fake reviews and ratings are also posted using automated bots, which can boost the visibility of those products on recommendation platforms \cite{kapoor2019corruption}. 
\begin{figure*}[t]
    \centering
    \includegraphics[width=0.95\linewidth]{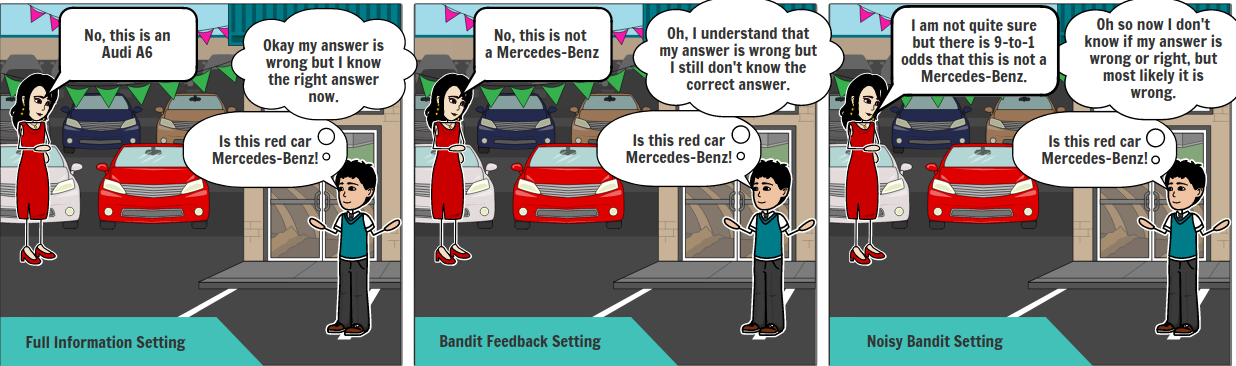}
    \caption{\footnotesize{Three kinds of supervised learning (a) Full Information Setting: In this setting, the learner receives the actual class label. (b) Bandit Feedback Setting: A bandit feedback is revealed to the learner, indicating whether the predicted label is correct or not. (c) Noisy Bandit Setting: The learner receives noisy bandit feedback (noisy feedback is received by flipping the correct feedback with some small probability).}}
    \label{fig:example}
\end{figure*}

In this paper, we model the noisy bandit feedback by assuming an adversary between the learner and the environment. Whenever the learner asks a binary query, the environment releases the actual feedback. Then, the adversary flips the 
actual feedback with probability $\rho$ and releases it to the learner.
The problem of multiclass classification under noisy bandit feedback is as follows: on each round, the learner is given an instance vector $\mathbf{x}$; the learner predicts a label $\hat{y}$; then the learner receives the corrupted feedback $f_{\rho}$. 
The noisy version of this problem is more challenging because, besides bandit feedback, the learner also has to deal with noise or corruption present in the feedback. 
To learn a robust classifier in the presence of noisy bandit feedback, we propose an unbiased estimator $h(f_{\rho})$ of the actual feedback $f$. The goal is to maximize the sum of $h(f_{\rho}^t)$, which in expectation, turns out to be the maximizing sum of actual feedbacks. Similar ideas have been explored to handle label noise in classification problems \cite{natarajan2013learning} under full information setting. This is the first work proposing a robust multiclass classifier under noisy bandit feedback to the best of our knowledge. 

\paragraph{{\bf Key Contribution of The Paper:}}
\begin{enumerate}
    \item We propose a robust algorithm for learning multiclass classifiers under noisy bandit feedbacks. The proposed algorithm enjoys a mistake bound of $O(\sqrt{T})$ in the high noise case and $O(T^{\nicefrac{2}{3}})$ in the worst case.
    \item We also propose an algorithm for noise rate estimation. 
    \item We validate our algorithms through experiments on benchmark datasets. 
\end{enumerate}


\section{Multiclass Classification} 
In the multiclass classification, the goal is to learn a function which maps each example to one of the $K$ categories. Let $g:\mathcal{X}\rightarrow [K]$ be the multiclass classifier where $\mathcal{X} \subseteq \mathbb{R}^d$ and $[K]=\{1,\ldots,K\}$. A multiclass classifier can be modeled using a weight matrix $W \in \mathbb{R}^{K\times d}$ as $g(\mathbf{x})={\arg\max}_{j\in [K]}\; \mathbf{w}_j\cdot\mathbf{x}$,
where $\mathbf{w}_j$ is the $j^{th}$ row of matrix $W$ and $\mathbf{x} \in {\cal X}$. We need to identify the weight matrix $W$ to find the classifier. In order to identify the parameters in $W$ of the underlying classifier, we use training data of the form $\{(\mathbf{x}^1,y^1),\ldots,(\mathbf{x}^T,y^T)\}$ where $(\mathbf{x}^t,y^t) \in \mathcal{X}\times \{1,\ldots,K\},\;\forall t\in [T]$. The performance of the classifier $f$ described by parameters $W$ on example $\mathbf{x}^t$ is measured using 0-1 loss as $L_{0-1}(g(\mathbf{x}^t),y^t)=\mathbb{I}[g(\mathbf{x}^t)\;\neq\; y^t]$.\footnote{Here, $\mathbb{I}[A]=1$ when the predicate $A$ is true and 0 otherwise.} $L_{0-1}$ is difficult to optimize. In practice, we use convex surrogates of $L_{0-1}$. $L_H$ is one such surrogate \cite{10.5555/944790.944813} described as follows.
\begin{align}
    L_H(W,(\mathbf{x}^t,y^t)) = \max_{j\neq y^t}\;[1-\mathbf{w}_{y^t}\cdot \mathbf{x}^t + \mathbf{w}_j\cdot\mathbf{x}^t]_+
    \label{eq:hingeloss}
\end{align}
Here $[a]_+=\max(0,a)$. Loss $L_H$ becomes 0 when $\mathbf{w}_{y^t}\cdot \mathbf{x}^t - \mathbf{w}_j\cdot \mathbf{x}^t \geq 1,\;\forall j\neq y^t$.

\subsection*{Online Multiclass Classification: Full Information Case}
In the full information case, the learner receives the actual class label of examples in every trial. A large margin Perceptron algorithm for multiclass classification using $L_H$ is proposed in \cite{10.1162/jmlr.2003.3.4-5.951}. The algorithm works as follows. The algorithm starts with $W^1$ as a zero matrix. Let $W^t$ be the weight matrix, and $\mathbf{x}^t$ be the example presented at trial $t$, to algorithm. Then the algorithm predicts the labels $\hat{y}^t$ as $\hat{y}^t={\arg\max}_{j\in [K]}\;\mathbf{w}_j^t\cdot \mathbf{x}^t$. Now it receives the true class label $y^t$ of $\mathbf{x}^t$. Algorithm incurs a loss $L_H(W^t,(\mathbf{x}^t,y^t))$ and updates the parameters as $W^{t+1}=W^t+U^t$. \begin{equation}
\label{eq:Perceptron_Update}
    U^t_{r,j}=\Big[\mathbb{I}[y^t=r]-\mathbb{I}[\hat{y}^t=r]\Big]x_{t,j}.
\end{equation}
This algorithm converges in finite iterations if the data is linearly separable \cite{10.1162/jmlr.2003.3.4-5.951}.

\subsection*{Online Multiclass Classification: Bandit Feedback Case}
In the bandit feedback setting \cite{kakade2008efficient}, the learner can only know whether the predicted label is correct or not. Banditron \cite{kakade2008efficient} modifies the Perceptron algorithm to deal with the bandit feedback. Let $W^t$ be the weight matrix in the beginning of trial $t$ and  $\mathbf{x}^t$ be the example presented at trial $t$. Let $\hat{y}^t = {\arg\max}_{j\in [K]}\;\mathbf{w}_j^t\cdot\mathbf{x}^t$. Banditron defines a probability distribution $p^t$ on class labels as follows.
\begin{align}
\label{eq:sample_dist}
    p^t(i)=(1-\gamma)\mathbb{I}\;[i=\hat{y}^t] +\frac{\gamma}{K}
\end{align}
Here, $\gamma \in [0,1)$ is the probability of exploration. The algorithm predicts the label $\tilde{y}^{t}$, which is randomly drawn from the distribution $p^{t}$. The algorithm then receives a feedback $f^t=\mathbb{I}[\tilde{y}^{t}=y^{t}]$. Banditron updates the weight matrix as $W^{t+1} = W^{t} + \tilde{U}^{t}$
where $
    \tilde{U}_{r,j}^{t} = x_{t,j} \displaystyle \left( \frac{\mathbb{I}[y^{t}=\tilde{y}^{t}] \mathbb{I}[\tilde{y}^{t} =r]}{p^t(r)} - \mathbb{I}[\hat{y}^{t} = r] \right)$.

\section{Learning Using Noisy Bandit Feedback}
In the noisy feedback setting, an adversary is present between the learner and the feedback, which manipulates the feedback to confuse the learner. It is hypothetical to assume noise-free data \cite{kapoor2019corruption} in the real world. So, one can find many real-world applications which are more appropriately modeled using a noisy feedback setting. For example, in a click-based recommendation system, we try to model the user behavior based on the clicks. These clicks are nothing but the bandit feedbacks, which are assumed to describe whether the user liked the recommended ad/product. Indeed, a user clicking the ad (or like the product) and likes it are two correlated events. But, the user may like the ad and does not click on it. 
On the other hand, the user may not like the ad but clicks on it (accidentally or in the absence of other exciting ads). These clicks are noisy as each user click does not necessarily mean that they agree with the recommended ad/product. 

In this paper, we model the noisy bandit feedback as follows. Let there be an adversary which flips the true feedback, $f$, with a non-zero probability and generates noisy feedback. We denote the noisy bandit feedback by $f_{\rho}$. Let $P(f_{\rho}=1|f=0)=\rho_{0}$, $P(f_{\rho}=0|f=1)=\rho_{1}$ be the noise rates
($\rho_{1} + \rho_{0} < 1$).

\subsection*{Proposed Approach}
Here, we propose a robust algorithm that can learn the true underlying classifier given noisy bandit feedback. To deal with the noisy or corrupted feedback, we propose a modified or proxy feedback $h(f_{\rho})$, which is an unbiased estimator of true feedback $f$, as follows. Given the noisy feedback $\f$, Lemma~1 shows how to construct an unbiased estimator of the true feedback $f$.\footnote{All the omitted proofs can be found in the supplementary material.}
\begin{lemma}
\label{lm:unbias_estimator}
Let $f^t=\mathbb{I}[\tilde{y}^{t}=y^{t}]$ be the true feedback. Let $h(f_\rho^t)$ be defined as,
\begin{equation}
 h(f_{\rho})= \frac{(1-\rho_{\f^{'}})\f - \rho_{\f}\f^{'}}{1-\rho_{0} - \rho_{1} }
 \label{eq:unbiased_estimator}
\end{equation}
where $\f^{'}=1-\f$. Then,
$\E_{f_{\rho}^t}[h(f_{\rho}^t)] = \mathbb{I}[\tilde{y}^{t}=y^{t}]=f^t.$
\end{lemma}

Instead of noisy feedback $f_\rho$, we use $h(f_\rho)$ (see eq~(\ref{eq:unbiased_estimator})) which is an unbiased estimator of the true feedback $f$ (Lemma~\ref{lm:unbias_estimator}). Similar ideas have been used to deal with the label noise in full information case \cite{natarajan2013learning}. 
We are now in a position to state a robust classifier for noisy bandit feedback. When there is no noise (\emph{i.e}, $\rho_0=\rho_1=0$), we see that $h(\f)=\f=f$. Thus, under noise-free case, $h(\f)$ becomes same as the noise-free bandit feedback $f$. At each round, the learner finds $\hat{y}^{t}=\arg\max_{j\in [K]}\;(\mathbf{w}_j^t\mathbf{x}^t)$ and defines a distribution $P^t$ over the class labels as described in eq~(\ref{eq:sample_dist}). Now, it samples a label $\tilde{y}^t$ randomly from $P^t$. It receives noisy bandit feedback $f_{\rho}^t$. We find $h(f_{\rho}^t)$ and update as $W^{t+1} = W^t + H^t$, where
\begin{equation}
    \label{eq:H_update}
    H^t_{r,j}=x^{t}_{j} \displaystyle \left(\frac{h(\f^t)\mathbb{I}[\tilde{y}^{t}=r]} {P^t(r)} - \mathbb{I}[\hat{y}^{t} = r] \right).
\end{equation}
$H^t$ has two sources of randomness, namely, $\tilde{y}^t$ (randomness used in the RCNBF algorithm) and $\f^t$ (randomness due to noise). Lemma~2 shows that the update matrix $H^t$ used in RCNBF is an unbiased estimator of the matrix $U^t$ (used in multiclass Perceptron), described in eq~(\ref{eq:Perceptron_Update}).
\begin{lemma}
\label{lm:update_matrix_estimator}
Suppose $H^{t}$ be the update matrix as defined in eq~(\ref{eq:H_update})  and let $U^{t}$ be the matrix as defined in eq~(\ref{eq:Perceptron_Update}). Then, $\E_{\tilde{y}^t,\f^t}[H^{t}]$ = $U^{t}$, where $\E_{\tilde{y}^t,\f^t}[H^{t}]$ is the expected value conditioned on $y^{1}, \cdots,y^{t-1}$.
\end{lemma}
We keep repeating these steps for $T$ trials. Complete details of the approach are given in Algorithm~\ref{alg:RCNBF}.
\subsection*{Mistake Bound Analysis of RCNBF}
In this section, we derive the mistake bound for the RCNBF (Algorithm~\ref{alg:RCNBF}). To do that, we first show that the expected value of the norm of $H^t$ is bounded.
\begin{lemma}
\label{lm:norm_matrix}
Let $H^{t}$ be defined as in eq~(\ref{eq:H_update}) and $\beta = 1-\rho_{0} -\rho_{1}$. Then,
\begin{align*}
    \E_{\tilde{y}^t,\f^t}[\norm{H^{t}}^{2}] \leq \displaystyle \norm{\mathbf{x}^{t}}^{2} \bigg( A_{1} \mathbb{I}[y^{t} \neq \hat{y}^{t}] +  A_2\mathbb{I}[y^{t} = \hat{y}^{t}] \bigg)
\end{align*}
where $A_1 = \frac{2K}{\gamma}  + \frac{2\rz(1-\rz)K}{\beta \gamma} + \frac{K\ri}{\beta^2\gamma} + \frac{\rz(1-\rz)K^2}{\beta^2\gamma^2}$, $A_2 = 2\gamma + \frac{\ri}{\beta^{2}(1-\gamma)} + \frac{\rz (1-\rz)K^2}{\beta^{2}\gamma}$.
\end{lemma}

\begin{center}
\begin{minipage}[t]{0.48\textwidth}
    \begin{algorithm}[H]
    \small{
    \caption{\underline{R}obust \underline{C}lassifier for  \underline{N}oisy  \underline{B}andit  \underline{F}eedback (RCNBF)}
    \label{alg:RCNBF}
    \textbf{Input}: $\gamma \in (0,0.5)$, $\rho_{0},\rho_{1} : \rho_{0} + \rho_{1} < 1 $\\
    \textbf{Initialize}: Set $W^{1} = 0 \in \mathbb{R}^{K \times d}$ 
    \begin{algorithmic} 
    \FOR{$t=1,2,\cdots,T$}
    \STATE Receive  $\mathbf{x}^{t} \in \mathbb{R}^{d}$. 
    \STATE Set $\hat{y}^{t} = \arg \max_{r\in[K]}(\mathbf{w}_r^{t}\cdot\mathbf{x}^{t})$
    \STATE Set $P^t(r) = (1-\gamma)\mathbb{I}[r=\hat{y}^{t}] + \tfrac{\gamma}{K},\;\forall r$
    \STATE Randomly sample $\tilde{y}^{t}$ according to $P^t$. 
    \STATE Predict $\tilde{y}^{t}$ and receive feedback $\f^t$
    \STATE Calculate $h(\f^t)$ using 
    \begin{equation*}
        h(f_{\rho}^t)= \tfrac{(1-\rho_{\f^{t'}})\f^t - \rho_{\f^t}\f^{t'}}{1-\rho_{0} - \rho_{1} }
    \end{equation*}
    \STATE Compute $H^{t} \in \mathbb{R}^{K \times d} $ such that
    \begin{equation*}
        H^{t}_{r,j} = x^{t}_{j} \displaystyle \left(\tfrac{h(\f^t)\mathbb{I}[\tilde{y}^{t}=r]}     {P^t(r)} - \mathbb{I}[\hat{y}^{t} = r] \right)
    \end{equation*}
    \STATE Update: $W^{t+1}$ = $W^{t} + H^{t}$
    \ENDFOR
    \end{algorithmic}
    }
    \end{algorithm}
\end{minipage}
\hfill
\begin{minipage}[t]{0.50\textwidth}
    \begin{algorithm}[H]
  \small{
    \caption{\underline{RC}NBF with \underline{I}mplicit \underline{N}oise \underline{E}stimation (RCINE)}
    \label{alg:RCINE}
    \textbf{Input}: $\gamma \in (0,0.5)$, $N_{s}$\\ 
    \textbf{Initialize}: $W^{1} = 0 \in \mathbb{R}^{K \times d}, \hat{\rho}_{0}=\hat{\rho}_{1}=0, \mathcal{S} $ 
    \begin{algorithmic} 
    \FOR{$t=1,2,\cdots,T$}
    \STATE Receive  $\mathbf{x}^{t} \in \mathbb{R}^{d}$. 
    \STATE Set $\hat{y}^{t} = \arg \max_{r\in[K]}(\mathbf{w}_r^{t}\cdot\mathbf{x}^{t})$
    \STATE Set $P^t(r) = (1-\gamma)\mathbb{I}[r=\hat{y}^{t}] + \tfrac{\gamma}{K},\;\forall r$
    \STATE Randomly sample $\tilde{y}^{t}$ according to $P^t$. 
    \STATE Predict $\tilde{y}^{t}$ and receive feedback $\f^t$ 
    \STATE Calulate $h(\f^t)$ using 
    
    \begin{equation*}
        h(f_{\rho}^t)= \tfrac{(1-\hat{\rho}_{\f^{t'}})\f^t - \hat{\rho}_{\f^t}\f^{t'}}{1-\hat{\rho}_{0} - \hat{\rho}_{1} }
    \end{equation*}
    \STATE Define $H^{t} \in \mathbb{R}^{K \times d} $ such that
    \begin{equation*}
        H^{t}_{r,j} = x^{t}_{j} \displaystyle \left(\tfrac{h(\f^t)\mathbb{I}[\tilde{y}^{t}=r]}     {P^t(r)} - \mathbb{I}[\hat{y}^{t} = r] \right)
    \end{equation*}
    \STATE Update: $W^{t+1}$ = $W^{t} + H^{t}$
    \STATE Data: Push $\{(\mathbf{x}^t,\tilde{y}^t),\f^t\} \; \text{in} \; \mathcal{S}$
    \IF{ $t\%N_s == 0$ }
        \STATE $\hat{\rho}_{0},\hat{\rho}_{1}$ = NREst($\mathcal{S}$), Clear $\mathcal{S}$
    \ENDIF
    \ENDFOR
    \end{algorithmic}
    }
    \end{algorithm}
    
\end{minipage}
\end{center}

Note that the norm of the matrix $H^t$ is inversely proportional to $\beta=1-\rho_0-\rho_1$. Thus, if noise rate increases, the upper bound on the norm of $H^t$ will increase. We now find expected mistake bound of the RCNBF algorithm.
\begin{theorem}[Mistake Bound]
Let $\mathbf{x}^{1}, \cdots ,\mathbf{x}^{T}$ be the sequence of examples presented to the RCNBF in $T$ trials. Let, $\norm{\mathbf{x}^{t}} \leq 1,\forall t\in [T] $ and $y^{t}\in [K]$. 
Let $R_{H}=\sum_{t=1}^{T} L_H(W^{*};(\mathbf{x}^{t},y^{t}))$ and $D=\norm{W^{*}}^{2}_{F} = \sum_{r=1}^{K}\sum_{j=1}^{d}(W_{i,j}^{*})^{2}$ be the cumulative hinge loss and the complexity of any matrix, $W^{*}$. Let $\rho_{0}$ and $\rho_{1}$ be the noise parameters. Then the expected number of mistakes made by RCNBF is upper bounded as
$
    \E[M] \leq R_{H}+ \sqrt{ A_1 DR_{H}} +  
    3 \max\big\{ A_1D, \sqrt{ A_2DT } \big\}
 + \gamma T$. Here, expectation is with respect to all the randomness of the algorithm.
\label{th:mistake bound}
\end{theorem}
Before moving, let us find the optimal value for the exploration-exploitation parameter $\gamma$ and the corresponding mistake bound. 

\begin{corollary}
(Zero Noise Case, $\rz=\ri=0$) In this case the mistake bound of \emph{RCNBF} is of the order O($\sqrt{T}$) which can be obtained by setting $\gamma = O( T^{\nicefrac{-1}{2}})$.
\end{corollary}

\begin{corollary}
(High Noise Case, $\rz,\ri \leq \min\big\{0.5 , O( \sqrt{\frac{D}{T}})\big\} $)
In this case,  we obtain the bound $\E[M] \leq O(\sqrt{DT}\beta^{-1})$ for $\gamma = O( \sqrt{\frac{D}{\beta^2 T}})$.
\end{corollary}

\begin{corollary}
(Very High Noise Case, $\rz,\ri \leq 1$) 
In this case the mistake bound of is $O(T^{\nicefrac{2}{3}}\beta^{-1})$ for $\gamma = O( {T}^{\nicefrac{-1}{3}}\beta^{-1})$.
\end{corollary}

We see that the above mistake bound is inversely proportional to $\beta$, \emph{i.e.}, as we increase the noise rate, the mistake bound will increase, which is as expected and also aligns with the batch mode algorithm in the presence of label noise \cite{natarajan2013learning}. 

\subsection*{Noise Rate Estimation}
Here, we propose an approach for estimating $\rho_0$ and $\rho_1$ which uses ideas presented in \cite{patrini2017making,liu2015classification}. The proposed approach is based on the following Theorem. 

\begin{theorem}
\label{th:noise_estimate}
Assume that 
\begin{enumerate}
    \item There exist at least one ``perfect example" for every class $j\in [K]$. Which means, there exists $\mathbf{x}_{j}^* \in \mathcal{X}$(prefect example for class j) such that $p(\mathbf{x}_{j}^*) > 0$ and $p(y=\tilde{y}|\mathbf{x}^*_{j},\tilde{y}=j) = p(y=j|\mathbf{x}_j^*)=1$.
    \item There exist sufficient corrupted examples to estimate $p(\f|\mathbf{x},\tilde{y}=l)$ accurately.
\end{enumerate}
Then it follows that $1-\rho_1 =  p(\f=1|\mathbf{x}_{l}^*,\tilde{y}=l),\;l\in[K]$  and $ \rho_0  = p(\f=1|\mathbf{x}^*_{k},\tilde{y}=l),\;l\neq k$, 
where $\mathbf{x}^*_l$ and $\mathbf{x}^*_k$ are perfect examples of class $l$ and $k$.
\end{theorem}
    \begin{algorithm}[t]
    \small{
    \caption{\underline{N}oise \underline{R}ate \underline{E}stimator (NREst)}
    \label{alg:noise_estimation}
    \textbf{Input:} $\mathcal{S}=\{\left(\mathbf{x}^t,\tilde{y}^t),\f^t\right)\;:\;t=1\ldots T\}$
    \begin{algorithmic} 
    \STATE Train a network using $\mathcal{S}$ which approximates $
        q(\mathbf{x},\tilde{y})=\hat{p}(\f=1|\mathbf{x},\tilde{y})$
    \STATE Find $
    \mathbf{x}^{j} = \arg\max_{\mathbf{x}\in \mathcal{X}} \; \hat{p}(\f=1|\mathbf{x},\tilde{y}=j),\;j\in [K]$
    \STATE Set $1-\rho_1 = \hat{p}(\f=1|\mathbf{x}^{l},\tilde{y}=l)$ and $ \rho_0 = \hat{p}(\f=1|\mathbf{x}^{k},\tilde{y}=l)$
    \end{algorithmic}
    \textbf{Output:} $\rho_{0},\rho_{1}$
    }
\end{algorithm}

Theorem~\ref{th:noise_estimate} assumes that for every class $j\in [K]$, there exists a perfect example $\mathbf{x}^*_j$ such that $p(f=1|\mathbf{x}^*_j,\tilde{y}=j)=p(y=j|\mathbf{x}^*_j)=1$. We use this idea to estimate the noise rates as follows. We use the data generated by RCNBF under noisy bandit feedback setting. Using this, we create a training set $\mathcal{S}$ with following sequence of examples $\{(\mathbf{x}^t,\tilde{y}^t),\f^t\}$ for $t=1\ldots N_{s}$. Note that the input to the network is $\mathbf{x}^t$ concatenated with $\tilde{y}^t$. This is the major difference with the noise rate estimation presented in \cite{patrini2017making}. We use $\mathcal{S}$ to train a neural network with a output layer of size 2 and softmax as the activation function of the output layer. Our classification problem is binary however following \cite{sivaprasad2020curious}, we prefer to use softmax with one-hot output instead of sigmoid as it allows the  network to learn non-convex boundaries. This network approximates $q(\mathbf{x},\tilde{y})=\hat{p}(\f=1|\mathbf{x},\tilde{y})$. Now we find perfect example for each class. A perfect example $\mathbf{x}^*_{j}$ for class $j$ is the one for which $\hat{p}(y=j|\mathbf{x}^*_{j})=\hat{p}(\f=1|\mathbf{x}^*_{j},\tilde{y}=j)=1$. We can find $\mathbf{x}^*_{j}$ as 
\begin{equation}
    \label{eq:perfect label}
    \mathbf{x}^*_{j} = \arg\max_{\mathbf{x}\in \mathcal{S}} \;  \hat{p}(\f=1|\mathbf{x},\tilde{y}=j),\;j\in [K]
\end{equation}
Now, we can approximate $\hat{\rho}_0$ and $\hat{\rho}_1$ as $
    1-\hat{\rho}_1 = \hat{p}(\f=1|\mathbf{x}^*_{l},\tilde{y}=l)$ and $\hat{\rho}_0= \hat{p}(\f=1|\mathbf{x}^*_{k},\tilde{y}=l)$.
The noise estimation approach is described in Algorithm~\ref{alg:noise_estimation}.
\begin{table*}[t]
\centering
\tiny{
  \caption{Estimated noise rates (rounded to 3 decimal digits)}
  \label{tb:estmate_noise_rates}
  \begin{tabular}{ll||ll|ll|ll}
    \toprule
    \multicolumn{2}{c||}{\multirow{2}{6em}{Actual Noise Rates}} & \multicolumn{6}{c}{ Estimated Noise Rates }\\
    \cmidrule(r){3-8}
    \multicolumn{2}{c||}{} & \multicolumn{2}{c|}{MNIST}  & \multicolumn{2}{c|}{USPS} & \multicolumn{2}{c}{Fashion-MNIST}\\
    \midrule
    \multicolumn{1}{c}{$\rho_{0}$} & \multicolumn{1}{c||}{$\rho_{1}$} & \multicolumn{1}{c}{$\;\;\;\hat{\rho}_{0}$} & \multicolumn{1}{c|}{$\;\;\;\hat{\rho}_{1}$} &
    \multicolumn{1}{c}{$\;\;\;\hat{\rho}_{0}$} & \multicolumn{1}{c|}{$\;\;\;\hat{\rho}_{1}$} &
    \multicolumn{1}{c}{$\;\;\;\hat{\rho}_{0}$} & \multicolumn{1}{c}{$\;\;\;\hat{\rho}_{1}$}\\
    \midrule
    0.000 &$\;$ 0.000 &$\;$ 0.063 &$\;$ 0.029   &$\;$ 0.017 &$\;$ 0.000  &$\;$ 0.090 &$\;$ 0.004\\
    0.150 &$\;$ 0.150 &$\;$ 0.172 &$\;$ 0.147   &$\;$ 0.181 &$\;$ 0.153  &$\;$ 0.189 &$\;$ 0.140\\
    0.250 &$\;$ 0.250 &$\;$ 0.248 &$\;$ 0.264   &$\;$ 0.258 &$\;$ 0.257  &$\;$ 0.264 &$\;$ 0.259\\
    0.200 &$\;$ 0.400 &$\;$ 0.211 &$\;$ 0.439   &$\;$ 0.194 &$\;$ 0.419  &$\;$ 0.215 &$\;$ 0.393\\
    0.400 &$\;$ 0.200 &$\;$ 0.400 &$\;$ 0.260   &$\;$ 0.393 &$\;$ 0.229  &$\;$ 0.404 &$\;$ 0.222\\
    0.400 &$\;$ 0.400 &$\;$ 0.403 &$\;$ 0.508   &$\;$ 0.402 &$\;$ 0.515  &$\;$ 0.397 &$\;$ 0.502\\
    \bottomrule
  \end{tabular}
 }
\end{table*}

\begin{figure*}[!tb]
    \centering
    \includegraphics[width=0.95\linewidth]{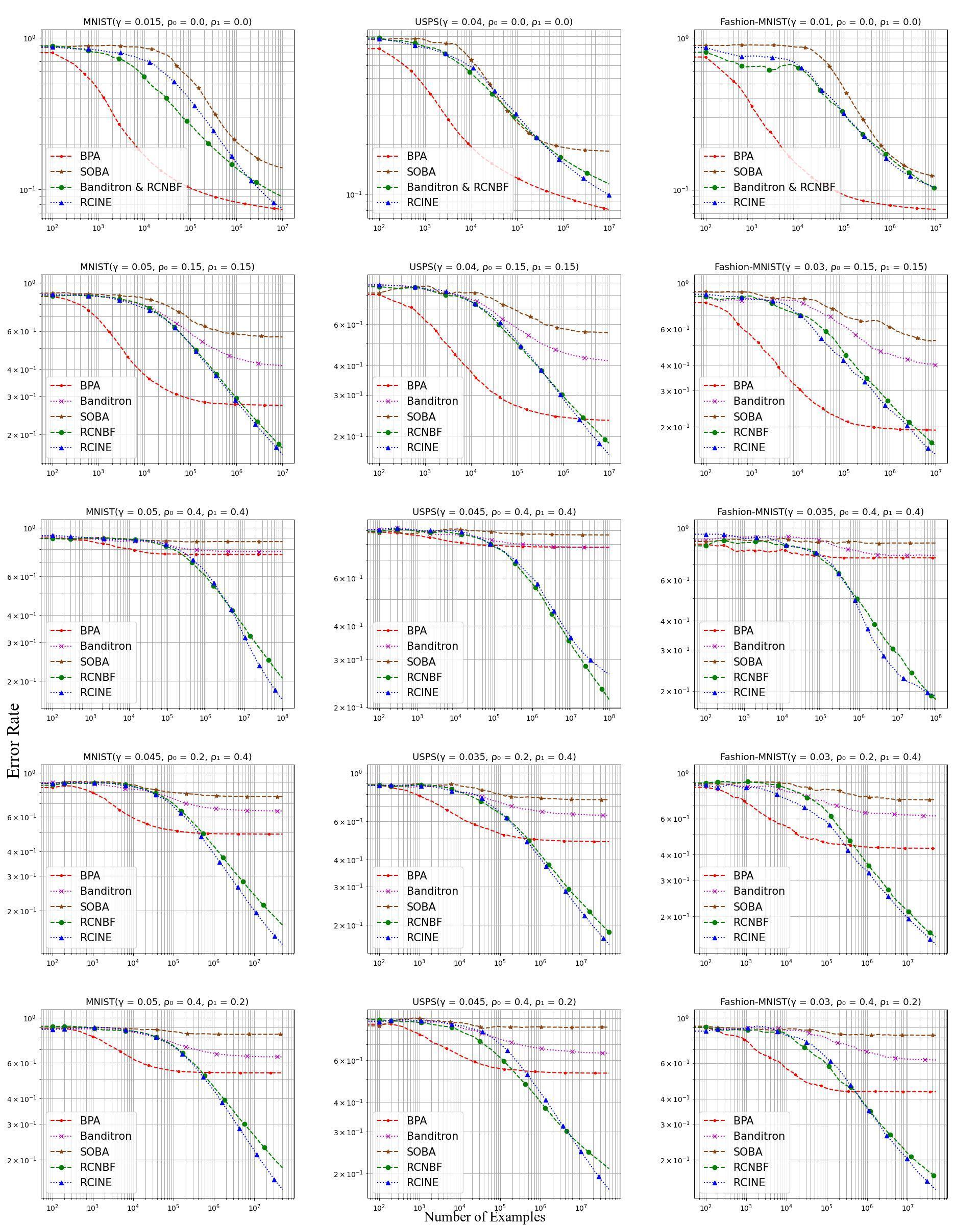}
    \caption{\footnotesize{Average error rates of RCNBF, RCINE and other benchmarking algorithms under noise-free case (first row; $\rho_{0}=\rho_{1}=0$), low noise case (second row; $\rho_{0}=\rho_{1}=0.15$), high noise case (third row; $\rho_{0}=\rho_{1}=0.40$) and mixed noise case (fourth row; $\rho_{0}=0.2, \rho_{1} =0.4$ and  fifth row; $\rho_{0}=0.4, \rho_{1} =0.2$). Three datasets are used (left to right): MNIST, USPS and  Fashion-MNIST.}}
    \label{fig:plot}
\end{figure*}
\begin{figure*}[t]
    \centering
    \includegraphics[width=0.95\linewidth]{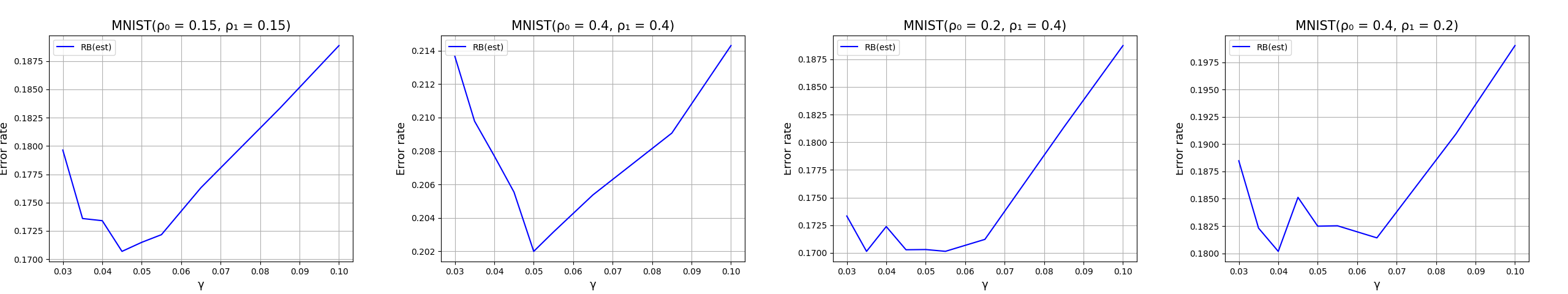}
    \caption{\footnotesize{Average error rates of RCINE against parameter's value $\gamma$ under different noise rate setting on MNIST.}}%
    \label{fig:gamma}
\end{figure*}

\subsection*{Learning using Noisy Bandit Feedback with Implicit Noise Rate Estimation}  
RCNBF (Algorithm~\ref{alg:RCNBF}) runs under the online setting while NREst (Algorithm~\ref{alg:noise_estimation}) is a batch algorithm. With the help of the above two algorithms, we are proposing a pseudo online mode algorithm, RCNBF with Implicit Noise Estimation(Algorithm~\ref{alg:RCINE}), which runs under the online setting. The RCINE Algorithm\footnote{The complete code for all the experiments can be found \href{https://github.com/Mudit-1999/Learning-Multiclass-Classifier-Under-Noisy-Bandit-Feedback-Code}{here}. } uses RCNBF to make predictions and generate dataset $\mathcal{S}$ for Noise Estimation. After every $N_{s}$ trails, the algorithm updates the estimated noise rate parameters by running the NREst algorithm on the collected dataset $\mathcal{S}$. The crux of this setup is that the RCNBF  will run in the online mode,  while NREst, which is running parallelly at the same time, will estimate the noise rates parameter $\hat{\rho}_{0}$  and $\hat{\rho}_{1}$  and update them repetitively after a small interval of time.

\section{Experimentation}
We do experiments on various real-world as well as synthetic datasets. The synthetic dataset is called SynSep. SynSep is a 9-class, 400-dimensional synthetic data set of size $10^{5}$. While constructing SynSep, we ensure that the dataset is linearly separable. For more detail about the dataset, one can refer to \cite{kakade2008efficient}. 
We also perform experiments on MNIST and Iris datasets from UCI repository \cite{Dua:2019}, USPS dataset\footnote{https://www.kaggle.com/bistaumanga/usps-dataset} 
and Fashion-MNIST for image classification \cite{xiao2017fashion}.
\footnote{The results and further discussion for SynSep and IRIS dataset are included in the supplementary file due to the space restrictions}. 

\paragraph{{\bf Feature Extraction for Fashion-MNIST dataset:}}
We first randomly sampled $35,000$ images from the dataset for feature extraction and trained a four-layer convolutional neural network. The first layer is a convolutional layer with 32 feature maps having a size of 3x3 and a stride of 1. It takes an input of 28 x 28 grayscale images. The convolutional layer is followed by a max-pooling layer having 2x2 as pool size. The next layer is a fully-connected layer with 100 units and a dropout of the probability of 0.2. The last layer is a fully connected softmax layer. To extract features, we took the output of the fully connected layer of size 100.  By experimenting on this dataset, we show that our approach can also be used for learning classifiers for complex datasets.

\paragraph{{\bf Benchmark Algorithms and Noise Rate Setting:}}
We present experimental comparisons of our proposed algorithms (RCNBF and RCINE) with Banditron \cite{kakade2008efficient}, Bandit Passive Aggressive \cite{zhong2015passive} and  Second Order Banditron Algorithm \cite{bmlc}. Five different settings of noise rate are used. 
These are (a) $\rho_{0}=\rho_{1}=0.0$, (b) $\rho_{0}=\rho_{1}=0.15$, (c) $\rho_{0}=\rho_{1}=0.4$, (d) $\rho_{0}=0.2,\rho_{1}=0.4$ and (e) $\rho_{0}=0.4, \rho_{1}=0.2$.  On each of the different noise setting, we ran our proposed algorithm, RCNBF (using original noise rates) and RCINE (with initial value of $\hat{\rho}_{0}=\hat{\rho}_{1}=0$). For updating the noise rates parameter, the RCINE algorithm, runs the NREst algorithm after $N_{s}$ trails on the collected dataset $\mathcal{S}$. NREst algorithm uses a neural network to estimate the noise rates. Table~\ref{tb:estmate_noise_rates} shows the results of estimation of noise rates at an intermediate instance of RCINE algorithm.

In NREst algorithm, train-test ratio of 90:10 is taken. Cross-entropy loss is chosen for comparison. 10$\%$ of the training set is used for validation. The mini-batch size used for training is 128. The activation function for all the network is ReLU and optimizer is AdaGrad \cite{duchi2011adaptive} with initial learning rate $0.01$ and $\delta = 10^{-6}$. After training, we apply the estimator to find $\hat{\rho}_{0}, 1-\hat{\rho}_{0}, \hat{\rho}_{1}$ and $1-\hat{\rho}_{1}$ on $\mathcal{S}$. Then we normalize the values of $\hat{\rho}_{0}, 1-\hat{\rho}_{0}$  and $\hat{\rho}_{1}, 1-\hat{\rho}_{1}$ such that they sum up to 1. From \cite{menon2015learning,patrini2017making} we know that the sample maximum is susceptible to the outliers, so instead of argmax eq~(\ref{eq:perfect label}), we take $89\%$-percentile.

For \textit{MNIST dataset}, the architecture consists of two dense hidden layers of size $128$ with a dropout of the probability of $0.2$. We train the network for 70 epochs. For the next set of experiments, we consider the \textit{USPS dataset}. We trained an architecture with three dense hidden layers of $32, 256$, and $32$ respectively, with a dropout of probability $0.2$ for 70 epochs. Lastly, for \textit{Fashion-MNIST dataset}, the architecture consists of three dense layers of size $32, 128$ and $32$ respectively with a dropout of probability $0.2$ and is trained for 70 epochs.

\paragraph{{\bf Parameter Selection:}} For each dataset and each different noise setting, simulations for RCINE are run for a wide range of values of the exploration parameter, $\gamma$. 
\footnote{ The value of $\gamma$ as shown in the figure are for RCINE. For other algorithms, the optimal value of $\gamma$ is chosen.}
For MNIST dataset, $\gamma$ exploration results are shown in Figure~\ref{fig:gamma}. We choose the $\gamma$ value for which the minimum error rate is achieved.

\paragraph{{\bf Results:}} We ran our proposed algorithms (RCNBF and RCINE) and compared the average \footnote{Note that here averaging is done over ten independent simulations of the algorithm} error rate with other benchmark algorithms as shown in Fig~\ref{fig:plot}. For better visualization of the asymptotic bounds, we plotted the result on a log-log scale. It shows that in the presence of noise, the final error rate of RCINE and RCNBF is significantly better than SOBA, BPA, and Banditron. While all other algorithms converge, RCNBF and RCINE are still learning and yet to converge. 


Analysis of Fig.~\ref{fig:plot} shows that as the number of examples grows, the slope of the error rate of RCNBF and RCINE under all different settings of noise rate is comparable to that of SOBA, BPA, and Banditron for the noise-free (0\%) setting. The final error rate of RCNBF and RCINE under all different noise rate settings is also close to SOBA, BPA, and Banditron under the noise-free setting. RCINE performs comparably to RCNBF for all the datasets and noise settings. This happens as we can efficiently estimate the noise rates. 

\section{Conclusion and Future Work}
In this paper, we proposed a noisy bandit feedback setting in online multiclass classification, which can effectively incorporate the noise present in real-world data. We proposed a novel algorithm based on the unbiased estimation technique, which enjoys a favorable bound (both theoretically and practically) under the proposed noisy bandit feedback setting. The proposed algorithm is robust to the noisy bandit feedback and can learn the true hypothesis in the presence of noise.  We also propose a technique to estimate the noise rate, thus providing an end-to-end framework. Experimental comparisons on various datasets with benchmarking algorithms show that RCNBF and RCINE are comparable to other algorithms under noise-free bandit feedback settings but far better than others under noisy bandit feedback settings. 

\bibliographystyle{plain}
\bibliography{bib_file}
\newpage
\appendix
\section*{Appendix}
\section{Proofs}
\subsection{Proof of Lemma~1}
\begin{equation}
\label{eq:unb_est}
    E_{f_{\rho}^t}[h(f_{\rho}^t)] = \mathbb{I}[\tilde{y}^{t}=y^{t}]
\end{equation}
We now consider two cases. 
\begin{enumerate}
    \item $\tilde{y}^{t} \neq y^{t}$: In this case, $\mathbb{I}[\tilde{y}^{t}=y^{t}]=0$. Now using the eq~(\ref{eq:unb_est}), we get
    \begin{equation}
    \label{eq:unb_est_case1}
    (1-\rho_{0})h(0) +  \rho_{0}h(1) = 0 
\end{equation}
\item $\tilde{y}^{t} = y^{t}$: In this case, $\mathbb{I}[\tilde{y}^{t}=y^{t}]=1$. Now using the eq~(\ref{eq:unb_est}), we get
\begin{equation}
\label{eq:unb_est_case2}
    (1-\rho_{1})h(1) +  \rho_{1}h(0) = 1
\end{equation}
\end{enumerate}  
Using equations (\ref{eq:unb_est_case1}) and (\ref{eq:unb_est_case2}) and solving for h(0) and h(1) gives
\begin{align*}
    h(0) = \frac{-\rho_{0}}{1-\rho_{0}-\rho_{1}}, \qquad \qquad
    h(1) = \frac{1-\rho_{0}}{1-\rho_{0}-\rho_{1}} 
\end{align*}
Combining these two equations, we can write 
\begin{equation*}
 h(f_{\rho})= \frac{(1-\rho_{\f^{'}})\f - \rho_{\f}\f^{'}}{1-\rho_{0} - \rho_{1} }
\end{equation*}
This concludes the proof.

\subsection{Proof of Lemma~2}
We see that
\begin{equation*}
    \E_{\tilde{y}^t,\f^t}[H^{t}] = \E_{\tilde{y}^t}[\E_{\f^t|\tilde{y}^t}[H^{t}|\tilde{y}^t]].
\end{equation*}
Now for each value of $r\in [K]$,$j\in [d]$, we have 
\begin{align*}
    &\E_{\tilde{y}^t,\f^t}[H^{t}_{r,j}] \\
    &=\displaystyle \E_{\tilde{y}^t}\left[\E_{\f^t|\tilde{y}^t}\left [x^{t}_{j}  \left(\frac {h(\f^t)\mathbb{I} [\tilde{y}^{t}=r]}{P^t(r)} - \mathbb{I}[\hat{y}^{t} = r] \right)|\tilde{y}^t\right]\right]\\
    &=\displaystyle \E_{\tilde{y}^t}\left[ x^{t}_{j} \left( \frac{ \E_{\f^t|\tilde{y}^t}[h(\f^t)|\tilde{y}^t] \mathbb{I} [\tilde{y}^{t}=r] }{P^t(r)} - \mathbb{I}[\hat{y}^{t} = r] \right)\right]
\end{align*}
Using Lemma 1, we get 
\begin{align*}
    &\E_{\tilde{y}^t,\f^t}[H^{t}_{r,j}]\\ 
    &=\displaystyle \E_{\tilde{y}^t}\left[x^{t}_{j}\left(\frac{ \mathbb{I}[\tilde{y}^{t}=y^{t}] \mathbb{I} [\tilde{y}^{t}=r] }{P^t(r)} - \mathbb{I}[\hat{y}^{t} = r] \right)\right] \\
    &= \displaystyle \sum_{i=1}^{k} P^t(i)x^{t}_{j}\left(\frac{ \mathbb{I}[i=y^{t}] \mathbb{I}[i=r]} {P^t(r)} - \mathbb{I}[\hat{y}^{t} = r] \right)\\
    &= \displaystyle x^{t}_{j}\left( \mathbb{I}[y^{t}=r]  - \mathbb{I}[\hat{y}^{t} = r] \right)=U_{r,j}^t
\end{align*}

\subsection{Proof of Lemma~3}
We observe that 
\begin{align*}
    \frac{\norm{H^{t}}^{2}}{\norm{\mathbf{x}^{t}}^{2}}\leq
 \begin{cases}
 \left(\frac{(1-\rho_{0})}{\beta P^t(y^{t})}\right)^{2}+1,\tilde{y}^{t}=y^{t}\neq\hat{y}^{t},\f=1 \\
 \left(\frac{\rho_{0}}{\beta P^t(y^{t})}\right)^{2}+1,\tilde{y}^{t}=y^{t}\neq\hat{y}^{t},\f=0 \\
 \left(\frac{(1-\rho_{0})}{\beta P^t(y^{t})}-1\right)^{2},\tilde{y}^{t}=y^{t}=\hat{y}^{t},\f=1 \\
 \left(\frac{\rho_{0}}{\beta P^t(y^{t})}+1\right)^{2},\tilde{y}^{t}=y^{t}=\hat{y}^{t},\f=0 \\
 \left(\frac{\rho_{0}}{\beta P^t(\tilde{y}^{t})}\right)^{2}+ 1,\tilde{y}^{t}\neq y^{t}=\hat{y}^{t},\f=0 \\
 \left(\frac{(1-\rho_{0})}{\beta P^t(\tilde{y}^{t})}\right)^{2}+1,\tilde{y}^{t}\neq y^{t}=\hat{y}^{t},\f=1 \\
 \left(\frac{\rho_{0}}{\beta P^t(\tilde{y}^{t})}+1\right)^{2},\tilde{y}^{t}\neq y^{t}, y^{t}\neq \hat{y}^{t},\f=0 \\
 \left(\frac{(1-\rho_{0})}{\beta P^t(\tilde{y}^{t})}\right)^{2}+1,\tilde{y}^{t}\neq y^{t}, y^{t}\neq \hat{y}^{t},\f=1 \\
 \end{cases}
    \end{align*}
Using towing property we get
\begin{align*}
    \frac{\E_{\tilde{y}^t,\rho}\big[\norm{H^{t}}^{2}\big]}{\norm{\mathbf{x}^{t}}^{2}} =  \frac{\E_{\tilde{y}^t}\big[\E_{\rho|\tilde{y}^t}[\norm{H^{t}}^{2}|\tilde{y}^t] \big]}{\norm{\mathbf{x}^{t}}^{2}}
\end{align*}
If $y^{t} = \hat{y}^{t}$, then
\begin{align*}
  &\frac{\E_{\tilde{y}^t,\f^t}\big[\norm{H^{t}}^{2}\big]}{\norm{\mathbf{x}^{t}}^{2}}\\
  =&P(\tilde{y}^t=y^{t})\frac{\E_{\f^t}[\norm{H^{t}}^{2}|\tilde{y}^t=y^t] }{\norm{\mathbf{x}^{t}}^{2}} + \big(1-P(\tilde{y}^t=y^{t})\big)\frac{\E_{\f^t}[\norm{H^{t}}^{2}|\tilde{y}^t\neq  y^{t}] }{\norm{\mathbf{x}^{t}}^{2}}\\
  =& P^t(y^{t})\bigg[(1-\rho_{1})\left( \frac{(1-\rho_{0})}{\beta P^t(y^{t})}-1\right)^{2} + \rho_{1} \left( \frac{\rho_{0}}{\beta P(y^{t})}+1\right)^{2} \bigg] \\ 
  &+ (1-P^t(y^{t}))\bigg[ (1-\rho_{0}) \left( \big(\frac{\rho_{0}}{\beta P^t(\tilde{y}^{t})}\big)^{2} + 1\right) + \rho_{0}\left( \big(\frac{(1-\rho_{0})}{\beta P^t(\tilde{y}^{t})}\big)^{2} + 1\right) \bigg]\\
  \leq&  P^t(y^{t}) \left[ 1 -\frac{2}{1-\gamma} + \frac{\beta^2+\ri}{\beta^2 (1-\gamma)^2}\right] + (1-P^t(y^{t}))\left[ 1+  \frac{ \rz (1-\rz)K^2}{\beta^{2}\gamma^{2}}\right]\\
  =& \frac{\gamma}{1-\gamma} + \frac{\ri}{\beta^{2}(1-\gamma)} + \frac{\rz (1-\rz)K^2}{\beta^{2}\gamma} \leq  2\gamma + \frac{\ri}{\beta^{2}(1-\gamma)} + \frac{\rz (1-\rz)K^2}{\beta^{2}\gamma} \\
\end{align*}

Similarly if $y^{t} \neq \hat{y}^{t}$, then
\begin{align*}
    \frac{\E_{\tilde{y}^t,\f^t}\big[\norm{H^{t}}^{2}\big]}{\norm{\mathbf{x}^{t}}^{2}} \leq & \frac{2K}{\gamma}  + \frac{2\rz(1-\rz)K}{\beta \gamma} + \frac{K\ri}{\beta^2\gamma} + \frac{\rz(1-\rz)K^2}{\beta^2\gamma^2} 
\end{align*}
Combining the two cases, we get the desired bound.

\subsection{Proof of Theorem 1}
We prove the theorem by evaluating the upper and lower bound of $\E[\langle W^{*},W^{T+1} \rangle]$, where $\langle W^{*},W^{t} \rangle$ := $\sum_{r=1}^{K}\sum_{j=1}^{d}W^{*}_{r,j}W^{t}_{r,j}$ 
Using the fact that $W^{1} = \mathbf{0}$, we can write $\E[\langle W^{*},W^{T+1} \rangle]$ as $\sum_{t=1}^{T} \Delta_{t}$ where 
\begin{equation*}
    \Delta_{t} :=  \E[\langle W^{*},W^{t+1} \rangle] - \E[\langle W^{*},W^{t} \rangle]
\end{equation*}
Using the definition of $W^{t+1}$ and Lemma 2, we get that for all $t$, we get
\begin{align*}
    \Delta_{t} = \E_{\mathbf{x}^t,y^t,\tilde{y}^t,\f^t}[\langle W^{*},H^{t}\rangle] = \E_{\mathbf{x}^t,y^t}[\langle W^{*},U^{t}\rangle] 
\end{align*}
Using the definition of hinge-loss as in eq (1), we can easily show that the following holds regardless of the value of $\hat{y}^{t}.$
\begin{equation*}
      L_H(W^{*},(\mathbf{x}^{t},y^{t})) \geq \mathbb{I}[\hat{y}^{t} \neq y^{t}] -  \langle W^{*},U^{t}\rangle
\end{equation*}
Rearranging the above equation and taking the expectation on both side,we get 
\begin{equation*}
      \Delta_{t} \geq \E[\mathbb{I}[\hat{y}^{t} \neq y^{t}]] - L_H(W^{*},(\mathbf{x}^{t},y^{t})) 
\end{equation*}
Summing over $t$, we obtain the lower bound
\begin{equation}
\label{eq:lower_bound}
    \E[\langle W^{*},W^{T+1} \rangle] = \sum_{t=1}^{T}\Delta_{t} \geq \E[\hat{M}] - R_{H} 
\end{equation}
where $\hat{M} := \sum_{t=1}^{T}\mathbb{I}[\hat{y}^{t} \neq y^{t}]$ and $R_{H}=\sum_{t=1}^{T} L_H(W^{*};(\mathbf{x}^{t},y^{t}))$ is the cumulative hinge-loss. Next we will evaluate the upper bound  of $\E[\langle W^{*},W^{T+1} \rangle]$. By using Cauchy-Schwartz inequality we get $\langle W^{*},W^{T+1} \rangle \leq \norm{W^{*}}* \norm{W^{T+1}}$. Let $D=\norm{W^{*}}^{2}_{F} = \sum_{r=1}^{K}\sum_{j=1}^{d}(W_{i,j}^{*})^{2}$. Exploiting the concavity of square root function and using Jensen's inequality, we obtain 
\begin{equation}
\label{eq:upper_bound}
    \E[\langle W^{*},W^{T+1} \rangle] \leq  \sqrt{D \E[\norm{W^{T+1}}^{2}]}
\end{equation}
Now, expanding $W^{T}$ by using the definition, we get 
\begin{align*}
    \E[\norm{W^{T+1}}^{2}] &= \E[\norm{W^{T}}^{2} + \langle W^{T},H^{T} \rangle + \norm{H^{T}}^{2} ]\\
    &= \sum_{t=1}^{T} \bigg( \E[\langle W^{T},H^{t} \rangle]  + \E[\norm{H^{T}}^{2}] \bigg)
\end{align*}
Using Lemma 2
we have for all $t$, $\E[\langle W^{*},H^{t}\rangle] = \E[\langle W^{*},U{t}\rangle] \leq 0 $, where the later inequality follows from the definition of $U^{t}$ and $\hat{y}^{t}$. Combining the above fact with Lemma 3 
and using the assumption $\norm{x^{t}} \leq 1$ for all $t$, we get 
\begin{align*}
    \E&[\norm{W^{T+1}}^{2}] \leq E\bigg[A_1\mathbb{I}[\hat{y}^{t} \neq y^{t}] +  A_2\mathbb{I}[\hat{y}^{t} = y^{t}]  \bigg]\\
\end{align*} 
where $A_1 = \frac{2K}{\gamma}  + \frac{2\rz(1-\rz)K}{\beta \gamma} + \frac{K\ri}{\beta^2\gamma} + \frac{\rz(1-\rz)K^2}{\beta^2\gamma^2}$ and $A_2 = 2\gamma + \frac{\ri}{\beta^{2}(1-\gamma)} + \frac{\rz (1-\rz)K^2}{\beta^{2}\gamma}$
\begin{align*}
    &\leq A_1 \E[\hat{M}] + A_2 T
\end{align*}
Substituting the above in the eq~(\ref{eq:upper_bound}) and using the inequality we obtain
\begin{align*}
    \E[&\langle W^{*},W^{T+1} \rangle] \leq \sqrt{ A_1D\E[\hat{M}]} + \sqrt{ A_2DT}
\end{align*}
Comparing the lower bound as given by eq~(\ref{eq:lower_bound}) with the above equation and reordering the terms
\begin{align*}
    \E[\hat{M}] &- \sqrt{ A_1D\E[\hat{M}]} - ( R_{H} + \sqrt{ A_2DT} ) \leq 0
\end{align*}
Standard algebraic manipulations give the bound
\begin{align*}
    \E[\hat{M}] \leq R_{H}+ \sqrt{ A_1 DR_{H}} +  
    3 \max\big\{ A_1D, \sqrt{ A_2DT } \big\}
\end{align*}
Since in expectation, we are exploring no more than $\gamma T$ of the rounds and thus $\E[M] \leq \E[\hat{M}] +\gamma T$
\subsubsection{Proof of Corollary 1}
\begin{align*}
    \E[M] \leq& R_H + \sqrt{\frac{2KDR_{H}}{\gamma}} + 3 \max\{\frac{2KD}{\gamma},\sqrt{2\gamma DT}\} + \gamma T
\end{align*}
Taking $\gamma = O(T^{\nicefrac{-1}{2}})$, we get
\begin{align*}
    =& O(T^{\nicefrac{1}{4}}) + 3 \max\{\sqrt{T},T^{\nicefrac{1}{4}}\} + O(\sqrt{T})\\
    =& O(\sqrt{T})\\
\end{align*}
\subsubsection{Proof of Corollary 2}
\begin{align*}
\text{Taking $\ri,\rz =$} &\text{$O(\sqrt{\frac{D}{T}})  \text{and} \gamma = O(\sqrt{\frac{D}{\beta^2T}}) $, we get}\\
    A_1 =& \frac{2K}{\gamma}  + \frac{2\rz(1-\rz)K}{\beta \gamma} + \frac{K\ri}{\beta^2\gamma} + \frac{\rz(1-\rz)K^2}{\beta^2\gamma^2}\\
    =&  \frac{2K\beta \sqrt{T}}{\sqrt{D}} + \frac{2K(\sqrt{T}-\sqrt{D})}{\sqrt{T}} + \frac{K}{\beta} + \frac{K^2(\sqrt{T}-\sqrt{D})}{\sqrt{D}}\\
    A_1D\leq& (2K\beta+K^{2})\sqrt{TD} + 2 K D + \frac{KD}{\beta}\\ 
    A_2 =& 2\gamma + \frac{\ri}{\beta^{2}(1-\gamma)} + \frac{\rz (1-\rz)K^2}{\beta^{2}\gamma}\\
    \leq & \frac{2\sqrt{D}}{\beta\sqrt{T}} + \frac{\sqrt{D}}{\beta^{2}\sqrt{T}} + \frac{K^2}{\beta}\\
    A_2DT \leq& (\frac{2}{\beta} + \frac{1}{\beta^2})D^{\nicefrac{3}{2}}\sqrt{T} +  \frac{K^2DT}{\beta}
\end{align*}
\begin{align*}
  \E&[M] \leq \sqrt{R_{H}\big((2K\beta+K^{2})\sqrt{TD} + 2 K D + \frac{KD}{\beta}\big)} \\
  &+ 3 \max\big\{ (2K\beta+K^{2})\sqrt{TD} + 2 K D + \frac{KD}{\beta}, \sqrt{(\frac{2}{\beta} + \frac{1}{\beta^2})D^{\nicefrac{3}{2}}\sqrt{T} +  \frac{K^2DT}{\beta}} \big\} \\  
  &+ \frac{\sqrt{DT}}{\beta}  \\
\end{align*}
Taking term with max power of T\\
\begin{align*}
    = O\big( \sqrt{DT}\beta^{-1} \big)
\end{align*}
\subsubsection{Proof of Corollary 3}
\begin{align*}
\text{Considering} &\;\ri,\rz \leq 1, \text{we get}\\
    A_1 \leq&  \frac{2K}{\gamma}  + \frac{2K}{\beta \gamma} + \frac{K}{\beta^2\gamma} + \frac{K^2}{\beta^2\gamma^2}\\
    A_2 \leq& 2\gamma + \frac{1}{\beta^{2}(1-\gamma)} + \frac{K^2}{\beta^{2}\gamma}
\end{align*}
Now taking $\gamma = O( {T}^{\nicefrac{-1}{3}}\beta^{-1})$, we get 
\begin{align*}
    A_1 =& O (2K{T}^{\nicefrac{1}{3}}\beta  + 2K{T}^{\nicefrac{1}{3}} + K{T}^{\nicefrac{1}{3}} + K^2{T}^{\nicefrac{2}{3}})\\
        =& O({T}^{\nicefrac{2}{3}})\\
    A_2 =& O\bigg(\frac{2{T}^{\nicefrac{-1}{3}}}{\beta} + \frac{1}{\beta^{2}} + \frac{K^2{T}^{\nicefrac{1}{3}}}{\beta}\bigg)\\
        =& O\bigg(\frac{K^2{T}^{\nicefrac{1}{3}}}{\beta^2}\bigg)\\
     \E[M] =& O\bigg( \sqrt{ {T}^{\nicefrac{2}{3}} DR_{H}} +  
    3 \max\big\{ {T}^{\nicefrac{2}{3}}D, \sqrt{ \frac{K^2{T}^{\nicefrac{4}{3}}}{\beta^2} } \big\}\bigg)\\
        =& O(\frac{T^{\nicefrac{2}{3}}}{\beta})
\end{align*}
\subsection{Proof of Theorem~2}
For any $\mathbf{x} \in \mathcal{X}$, we have
\begin{align*}
    p(\f=1|\mathbf{x},\tilde{y}=l)
    =&p(\f=1|f=1) p(f=1|\mathbf{x},\tilde{y}=l)\\
    &+p(\f=1|f=0)p(f=0|\mathbf{x},\tilde{y}=l)\\
    =(1-\rho_1)p(f&=1|\mathbf{x},\tilde{y}=l) + \rho_0 p(f=0|\mathbf{x},\tilde{y}=l)
\end{align*}
Using $f=\I[y=\tilde{y}]$, we get
\begin{align*} 
    p(\f=1|\mathbf{x},\tilde{y}=l) =& (1-\rho_1)p(\I[y=\tilde{y}]=1|\mathbf{x},\tilde{y}=l)\\ 
    &+\rho_0 p(\I[y=\tilde{y}]=0|\mathbf{x},\tilde{y}=l)
\end{align*}
\begin{align*}
    & =(1-\rho_1)p(y=\tilde{y}|\mathbf{x},\tilde{y}=l) + \rho_0 p(y\neq \tilde{y}|\mathbf{x},\tilde{y}=l)\\
    & =(1-\rho_1)p(y=l|\mathbf{x}) + \rho_0 p(y\neq l|\mathbf{x}).
\end{align*}
\begin{enumerate}
\item Now by assumption (a), when $\mathbf{x} = \mathbf{x}^{l}$ (perfect example for class $l$), $p(y=l|\mathbf{x}^{l})=p(y=\tilde{y}|\mathbf{x}^l,\tilde{y}=l) =1$ and $p(y=k|\mathbf{x}^{l})=0,\;k\neq l$.
Thus, 
\begin{align*}
    p(\f=1|\mathbf{x}^l,\tilde{y}=l) = 1-\rho_1
\end{align*}
\item When $\mathbf{x} = \mathbf{x}^{k}$ (perfect example of class $k$),
$p(y=l|\mathbf{x}^{k})=p(y=\tilde{y}|\mathbf{x}^k,\tilde{y}=l) =0$ and $p(y\neq l|\mathbf{x}^{k})=1,\;l\neq k$. Thus, we get
\begin{align*}
p(\f=1|\mathbf{x}^k,\tilde{y}=l) = \rho_0
\end{align*}
\end{enumerate}
And from (b), we can say the empirical estimate $\hat{p}(\f=j|\mathbf{x},\tilde{y}=l)$ is same as the true probability distribution $p(\f=j|\mathbf{x},\tilde{y}=l)$.

\section{Other Simulation Results}
Noise rate estimation results at an intermediate instance of RCINE for the Iris and SynSep dataset are shown in  Table~\ref{tb:appendix_estmate_nr}. For \textit{Iris dataset}, the architecture consists of two dense hidden layers of size $32$ with dropout probability 0.2 and is trained for 70 epochs. For the synthetic dataset,\textit{SynSep} the architecture consists of two dense hidden layers of size 48 with probability 0.3 of dropout and is trained for 70 epochs. Only for \textit{SynSep}, the estimator performs sub-optimal when we take $89\%$-percentile instead of argmax in eq(8), probably due to the synthetic nature of the dataset. So for estimating, we take  $94\%$ percentile instead of $89\%$.

\begin{table*}[t]
\centering
\footnotesize{
  \caption{Estimated noise rates (rounded to 3 decimal digits)}
  \label{tb:appendix_estmate_nr}
  \begin{tabular}{ll||ll|ll}
    \toprule
    \multicolumn{2}{c||}{\multirow{2}{6em}{Actual Noise Rates}} & \multicolumn{4}{c}{ Estimated Noise Rates }\\
    \cmidrule(r){3-6}
    \multicolumn{2}{c||}{} & \multicolumn{2}{c|}{IRIS} & \multicolumn{2}{c}{SynSep}\\
    \midrule
    \multicolumn{1}{c}{$\rho_{0}$} & \multicolumn{1}{c||}{$\rho_{1}$} & \multicolumn{1}{c}{$\hat{\rho}_{0}$} & \multicolumn{1}{c|}{$\hat{\rho}_{1}$} &
    \multicolumn{1}{c}{$\hat{\rho}_{0}$} & \multicolumn{1}{c}{$\hat{\rho}_{1}$}\\
    \midrule
    0.000 & 0.000 & 0.000 & 0.000   & 0.000 & 0.003\\
    0.150 & 0.150 & 0.164 & 0.172   & 0.177 & 0.143\\
    0.250 & 0.250 & 0.261 & 0.288   & 0.286 & 0.282\\
    0.200 & 0.400 & 0.209 & 0.423   & 0.230 & 0.459\\
    0.400 & 0.200 & 0.407 & 0.244   & 0.453 & 0.268\\
    0.400 & 0.400 & 0.412 & 0.412   & 0.416 & 0.499\\
    \bottomrule
    \bottomrule
  \end{tabular}
 }
\end{table*}

\begin{figure*}[!ht]
    \centering
    \subfloat
    []
    {{\includegraphics[width=0.30\linewidth]{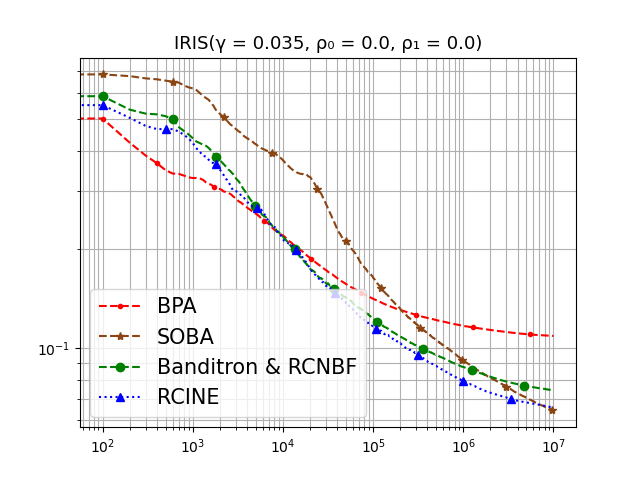} }}%
    \subfloat
    []
    {{\includegraphics[width=0.30\linewidth]{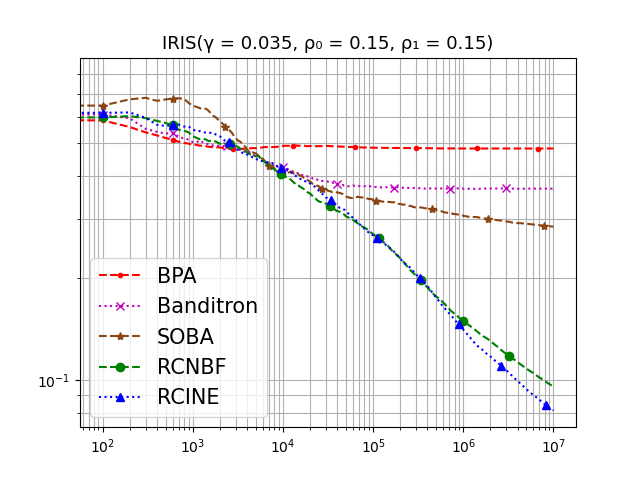} }}%
    \subfloat
    []
    {{\includegraphics[width=0.30\linewidth]{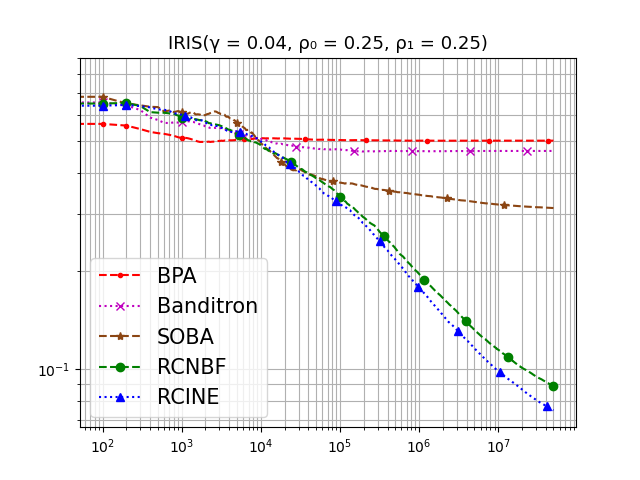} }}%
    \newline
    \subfloat
    []
    {{\includegraphics[width=0.30\linewidth]{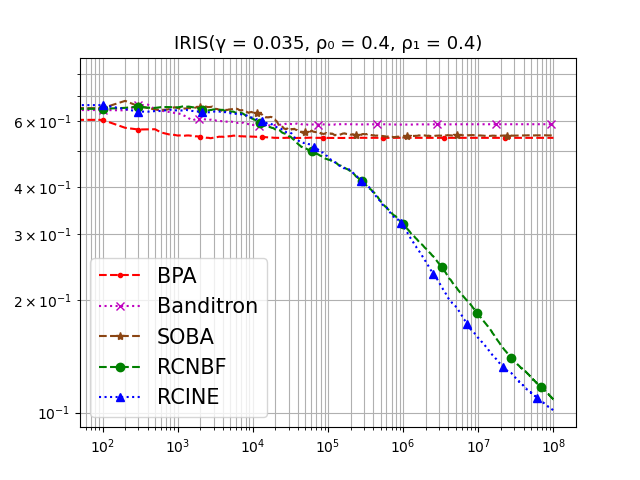} }}%
    \subfloat
    []
    {{\includegraphics[width=0.30\linewidth]{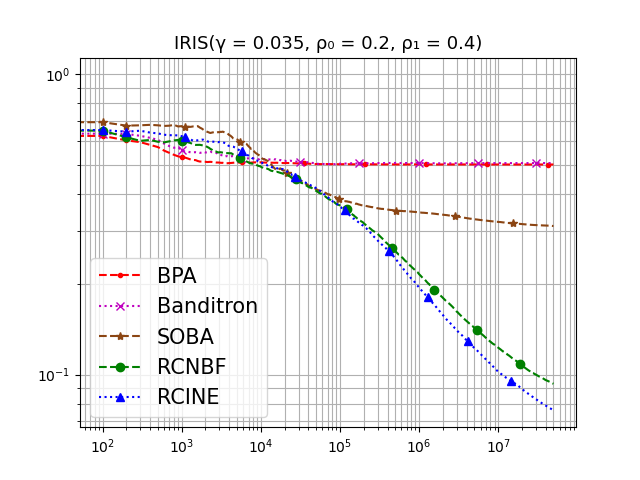} }}%
    \subfloat
    []
    {{\includegraphics[width=0.30\linewidth]{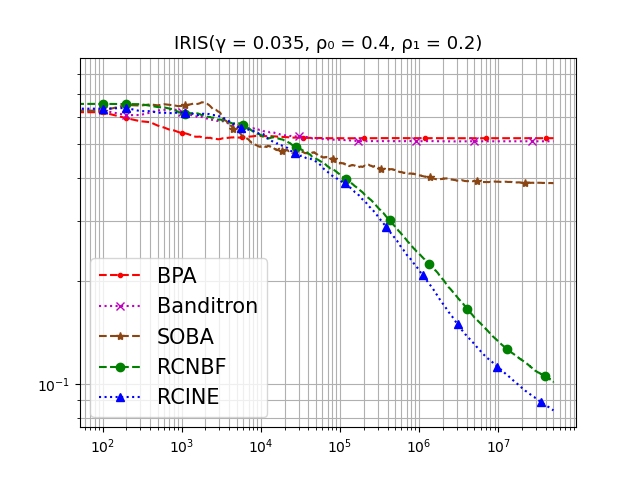} }}%
    \caption{Average error rates of RCNBF, RCINE and other benchmarking algorithms under (a) noise free case: $\rho_{0}=\rho_{1}=0$, (b) low noise case: $\rho_{0}=\rho_{1}=0.15$, (c) intermediate noise case: $\rho_{0}=\rho_{1}=0.25$, (d) high noise case: $\rho_{0}=\rho_{1}=0.40$, (e,f) mixed noise case: $\rho_{0}=0.2, \rho_{1} =0.4$ and $\rho_{0}=0.4, \rho_{1} =0.2$ for Iris dataset.}%
    \label{fig:iris_plot}
\end{figure*}

\begin{figure*}[!ht]
    \centering
    \subfloat
    []
    {{\includegraphics[width=0.30\linewidth]{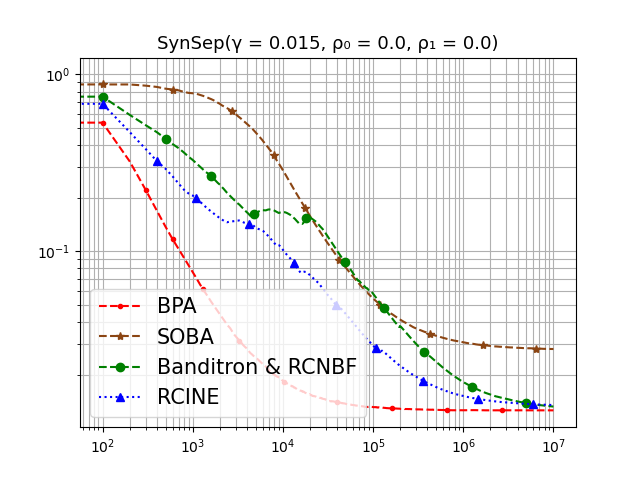} }}%
    \subfloat
    []
    {{\includegraphics[width=0.30\linewidth]{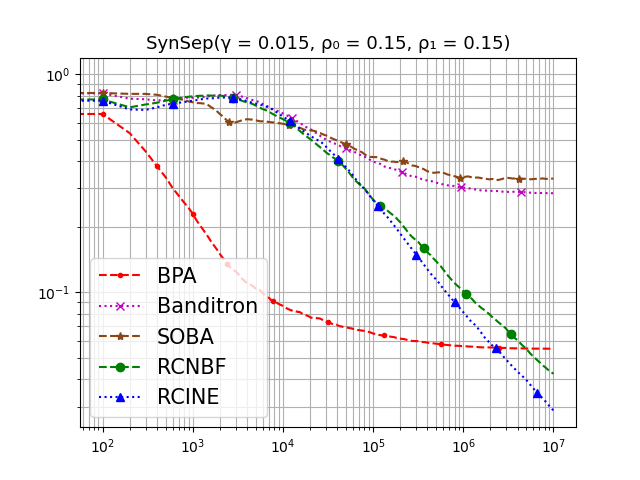} }}%
    \subfloat
    []
    {{\includegraphics[width=0.30\linewidth]{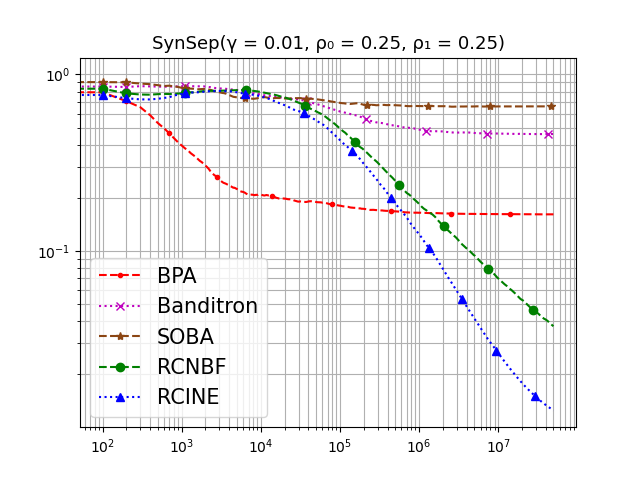} }}%
    \newline
    \subfloat
    []
    {{\includegraphics[width=0.30\linewidth]{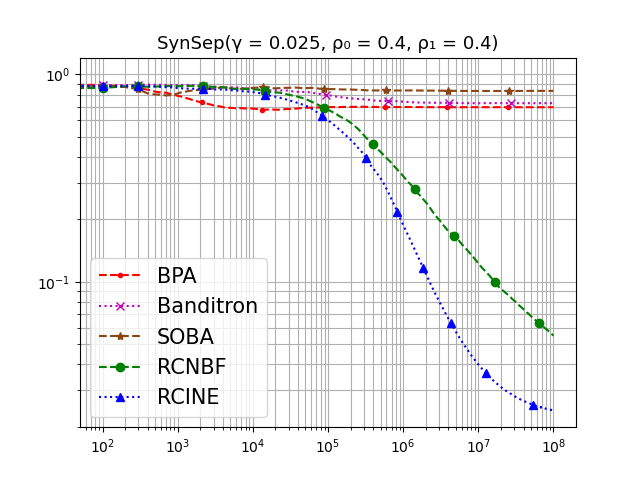} }}%
    \subfloat
    []
    {{\includegraphics[width=0.30\linewidth]{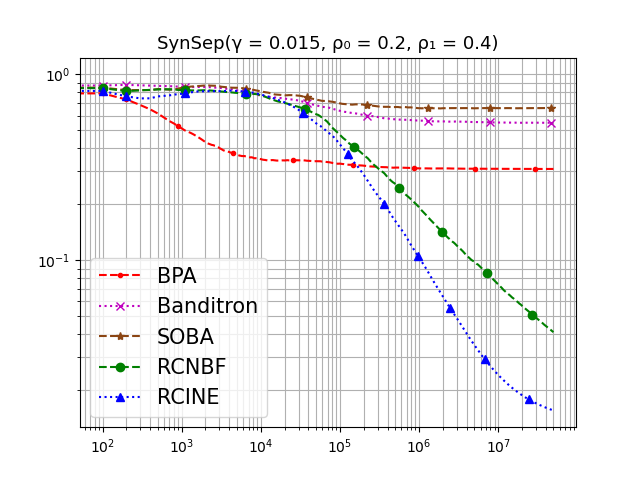} }}%
    \subfloat
    []
    {{\includegraphics[width=0.30\linewidth]{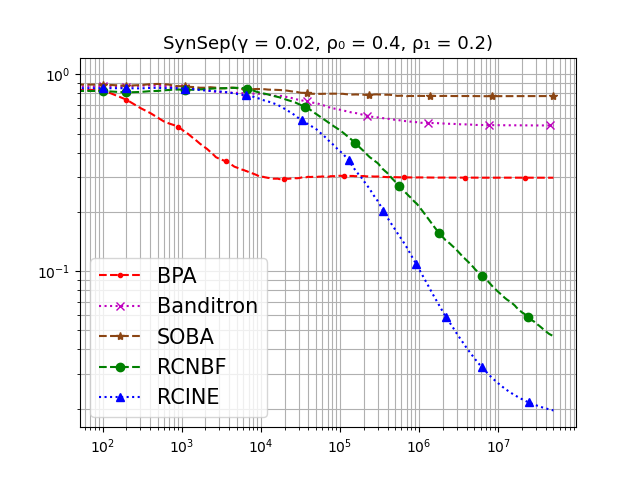} }}%
    \caption{Average error rates of RCNBF, RCINE and other benchmarking algorithms under (a) noise free case: $\rho_{0}=\rho_{1}=0$, (b) low noise case: $\rho_{0}=\rho_{1}=0.15$, (c) intermediate noise case: $\rho_{0}=\rho_{1}=0.25$, (d) high noise case: $\rho_{0}=\rho_{1}=0.40$, (e,f) mixed noise case: $\rho_{0}=0.2, \rho_{1} =0.4$ and $\rho_{0}=0.4, \rho_{1} =0.2$ on SynSep dataset.}%
    \label{fig:synsep_plot}
\end{figure*}

\subsection{Intermediate Noise Rates}
\begin{figure*}[!ht]
    \centering
    \subfloat
    {{\includegraphics[width=0.30\linewidth]{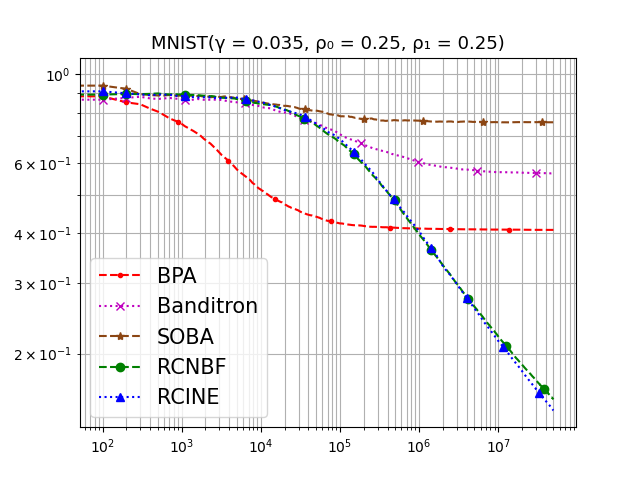} }}%
    \subfloat
    {{\includegraphics[width=0.30\linewidth]{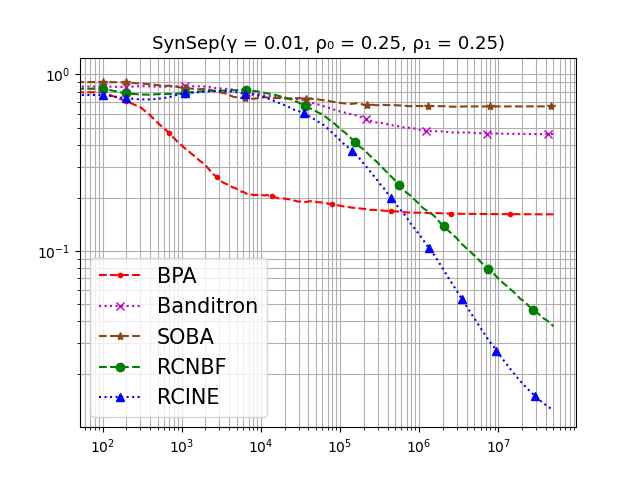} }}%
    \subfloat
    {{\includegraphics[width=0.30\linewidth]{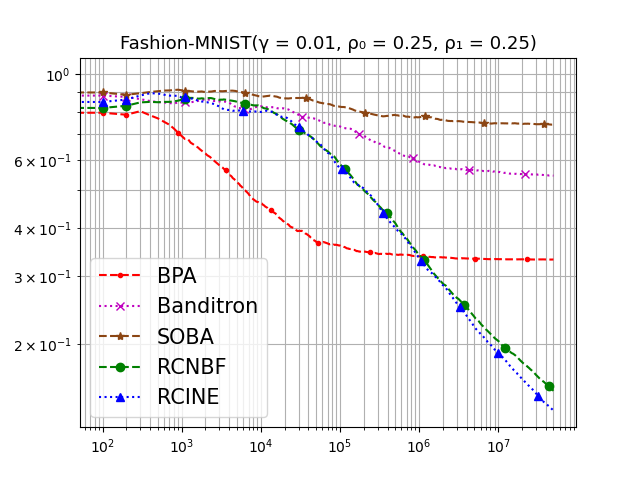}}}%
    \caption{Average error rates of RCNBF, RCINE and other benchmarking algorithms under  intermediate noise case ($\rho_{0}=\rho_{1}=0.25$) for MNIST, USPS and Fashion-MNIST datasets respectively.}
    \label{fig:intermediate_noise_rate}
\end{figure*}

Fig.~\ref{fig:intermediate_noise_rate} shows the plot for intermediate noise case ,\emph{i.e}, $\rz=\ri=0.25$ for MNIST, USPS and Fashion-MNIST dataset respectively.

\end{document}